\documentclass[10pt,twocolumn,letterpaper]{article}

\usepackage[pagenumbers]{cvpr} 

\usepackage{graphicx}
\usepackage{amsmath}
\usepackage{amssymb}
\usepackage{booktabs}
\usepackage{multirow}
\usepackage[accsupp]{axessibility}
\usepackage[pagebackref,breaklinks,colorlinks]{hyperref}

\usepackage[capitalize]{cleveref}
\crefname{section}{Sec.}{Secs.}
\Crefname{section}{Section}{Sections}
\Crefname{table}{Table}{Tables}
\crefname{table}{Tab.}{Tabs.}

\def\cvprPaperID{44} 
\def\confName{CVPR}
\def\confYear{2023}

\begin{document}

\title{Envisioning a Next Generation Extended Reality Conferencing System with Efficient Photorealistic Human Rendering}
\author{
Chuanyue Shen$^1$\thanks{Corresponding Author.}~\thanks{This work was done during an internship at Futurewei Technologies.}~~~~~
Letian Zhang$^2$\footnotemark[2]~~~~~
Zhangsihao Yang$^3$\footnotemark[2]~~~~~
\\
Masood Mortazavi$^4$~~~~~
Xiyun Song$^4$~~~~~
Liang Peng$^4$~~~~~
Heather Yu$^4$
\smallskip 
\\
$^1$University of Illinois at Urbana-Champaign~~~~~~~
$^2$University of Miami
\\
$^3$Arizona State University~~~~~~~
$^4$Futurewei Technologies
\\
{\small
$^1$cs11@illinois.edu~~
$^2$lxz437@miami.edu~~
$^3$zhangsihao.yang@asu.edu~~
$^4$\{masood.mortazavi, xsong, lpeng, hyu\}@futurewei.com
}}
\maketitle

\begin{abstract}
   Meeting online is becoming the new normal. Creating an immersive experience for online meetings is a necessity towards more diverse and seamless environments. Efficient photorealistic rendering of human 3D dynamics is the core of immersive meetings. Current popular applications achieve real-time conferencing but fall short in delivering photorealistic human dynamics, either due to limited 2D space or the use of avatars that lack realistic interactions between participants.
   Recent advances in neural rendering, such as the Neural Radiance Field (NeRF), offer the potential for greater realism in metaverse meetings. However, the slow rendering speed of NeRF poses challenges for real-time conferencing. 
   We envision a pipeline for a future extended reality metaverse conferencing system that leverages monocular video acquisition and free-viewpoint synthesis to enhance data and hardware efficiency. Towards an immersive conferencing experience, we explore an accelerated NeRF-based free-viewpoint synthesis algorithm for rendering photorealistic human dynamics more efficiently. We show that our algorithm achieves comparable rendering quality while performing training and inference $44.5\%$ and $213\%$ faster than state-of-the-art methods, respectively. Our exploration provides a design basis for constructing metaverse conferencing systems that can handle complex application scenarios, including dynamic scene relighting with customized themes and multi-user conferencing that harmonizes real-world people into an extended world.
\end{abstract}

\section{Introduction}
\label{sec:intro}
Meeting online is becoming the new normal. Typical scenarios include video conferencing, teamwork, and socializing when people are separated. Creating an immersive experience for online meetings could revolutionize industries such as business, education, and entertainment by enabling more efficient meetings, facilitating remote learning, and providing more diverse, capable, and pleasing environments. Currently, the most popular conferencing systems, such as Zoom, Teams, and Gather, offer a delightful audio and visual experience in 2D spaces. However, they do not or only weakly incorporate human dynamics from real world into 3D virtual space. Horizon by Meta creates a 3D virtual world, but it only animates real humans as half-body avatars in pre-assigned positions. These systems have limited realism, losing the realistic interaction between people and thus impoverishing the meeting experience. 

In realizing an immersive experience, a metaverse conferencing system needs to fulfil efficient photorealistic rendering based on free-viewpoint synthesis using acquired human full-body motion. Previous free-viewpoint rendering systems rely on acquiring motion videos from a group of densely arranged cameras \cite{Lumigraph, HPPFDB18}, or depth cameras \cite{collet2015high, dou2016fusion4d}. Recent advances allow the use of sparse multi-view video acquisition \cite{peng2021neural}. However, these systems all require expensive setup and maintenance, reducing the efficiency and the accessibility to many applications. A single-camera video acquisition is ideal for a system that benefits from enhanced efficiency and increased accessibility.  

In terms of rendering human dynamics, previous digital human research enables the rendering of avatars which simulate the human body language as in modeled artificial characters \cite{Lee2004Avatar}. Avatars are fun to use but they do not reveal the realistic appearance of human beings and thus are less engaging in online meeting scenarios. The introduction of NeRF brings new outlooks in synthesizing photorealistic novel views. It catalyzes a wave of human neural rendering methods that deliver high fidelity results \cite{peng2021neural, liu2021neuralactor, peng2021animatable, h-nerf, weng_humannerf_2022_cvpr, jiang2022neuman}. However, one drawback of NeRF is its slow training and rendering speed. Several works focus on accelerating NeRF \cite{liu2020nsvf, yu2021plenoctrees, garbin2021fastnerf, mueller2022instant, Reiser2021kilonerf, lindell2021autoint}. However, little effort has been done for accelerating human neural rendering. 

In this work, we envision a pipeline for extended reality conferencing and explore a more efficient human dynamic rendering algorithm based on NeRF. Our envisioned pipeline allows increased efficiency in data, hardware, and rendering, benefiting from a single-view video acquisition protocol and accelerated free-viewpoint synthesis. Specifically, our human dynamic rendering algorithm achieves comparable rendering quality while performing training and inference $44.5\%$ and $213\%$ faster than state-of-the-art methods, respectively. Our exploration provides insights for building future metaverse conferencing systems that offer immersive and real-time photorealistic experience.

\section{Background}
The core of our envisioned metaverse conferencing pipeline is NeRF-based free-viewpoint rendering of human dynamics. We review related background in this section.

\subsection{Neural Radiance Field}
\label{subsec: background_nerf}
Neural Radiance Field attracts tremendous attention in the fields of computer graphics, vision, and multimedia since its first introduction in 2020 by \cite{mildenhall2020nerf}. NeRF represents a scene from a set of multi-view images as a radiance field, and renders novel views of the scene from the radiance field. The view synthesis of NeRF obtains nearly more-than-ever photorealistic quality while the theory behind is rather simple given by \cref{eq:nerf_1}: input a 5D coordinate (including a 3D location $\textbf{x}$ and 2D viewing direction $\textbf{d}$) into a MLP $F_{\Theta}$ and output a volume density $\sigma$ and view-dependent RGB color $\textbf{c}$ at the corresponding location. 
\begin{equation}
    F_{\Theta}: (\textbf{x}, \textbf{d}) \rightarrow (\textbf{c}, \sigma)
    \label{eq:nerf_1}
\end{equation}

NeRF uses volume rendering to produce novel view images from output color $\textbf{c}$ and density $\sigma$. Based on classical volume rendering principles, the vanilla NeRF \cite{mildenhall2020nerf} composites color $C(\textbf{r})$ of camera ray $\textbf{r}(t) = \textbf{o} + t\textbf{d}$ via \cref{eq:nerf_2}:
\begin{equation}
    C(\textbf{r}) = \int_{t_n}^{t_f} T(t)\sigma(\textbf{r}(t))\textbf{c}(\textbf{r}(t), \textbf{d})\,dt
    \label{eq:nerf_2}
\end{equation}
where $T(t)$ is the accumulated transmittance along the ray from $t_n$ to $t$ given by $T(t) = \exp(-\int_{t_n}^{t} \sigma(\textbf{r}(s)\,ds))$.
With quadrature rule and stratified sampling approach \cite{mildenhall2020nerf}, $C(\textbf{r})$ can be numerically estimated as 
\begin{equation}
    \hat{C}(\textbf{r}) = \sum_{i=1}^{N} T_i(1-\exp{(-\sigma_i\delta_i)}\textbf{c}_i)
    \label{eq:nerf_3}
\end{equation}
where $\delta_i = t_{i+1} - t_i$ is the distance between adjacent samples, and $T_i = \exp(-\sum_{j=1}^{i-1} \sigma_j\delta_j$).
The vanilla NeRF \cite{mildenhall2020nerf} minimizes the loss, which is defined as the total squared error between the rendered and true pixel colors.

To optimize the neural network $F_{\Theta}$ for better fitting, the vanilla NeRF \cite{mildenhall2020nerf} uses positional encoding to project the 5D input to a higher dimensional space before passing into the MLP. To increase the rendering efficiency, the vanilla NeRF \cite{mildenhall2020nerf} proposes a hierarchical volume sampling method by implementing a coarse network and a fine network, where the former informs the latter to obtain sampling points of higher importance. These optimization strategies are adopted in later extended works of NeRF. 

Due to its high-quality performance while being simple and extendable, the use of NeRF as a core algorithm has been widely explored in a variety of scene representation and rendering tasks, such as pose estimation \cite{peng2021neural, peng2021animatable, su2021anerf, weng_humannerf_2022_cvpr}, lighting \cite{zhang2021nerfactor, boss2021nerd, chen2021nerv}, scene labeling and understanding \cite{Zhi:etal:ICCV2021, tancik2022blocknerf}, and scene composition \cite{Niemeyer2020GIRAFFE, yang2021objectnerf}. 

\subsection{NeRF for human}
Human is composed of a rigid skeleton and soft tissues. Human motion can be very articulated and causes the deformation of human body, thus imposing challenges in human reconstruction, animation, and rendering tasks. Previous human free-viewpoint rendering systems rely on a group of densely-spaced cameras \cite{Lumigraph, HPPFDB18}. Recent works attempt to realize free-viewpoint rendering with reduced hardware complexity, such as using videos acquired by sparse multi-view \cite{peng2021neural} or even monocular camera systems \cite{weng_humannerf_2022_cvpr, su2021anerf, jiang2022neuman}. 

NeRF stimulates the development of a wave of NeRF-based human animation and rendering methods. Neural Body \cite{peng2021neural} achieves free-viewpoint synthesis from a sparse multi-view video by leveraging the arts from the NeRF model, a Skinned Multi-Person Linear (SMPL) model \cite{SMPL:2015}, and a latent variable model \cite{loehlin2004latent}. By learning a set of latent codes anchored to a deformable mesh from SMPL, Neural Body generates novel views of the human subject at different poses. On top of Neural Body, Neural Human Performer \cite{kwon2021neural} enhances the rendering quality of unseen identities and poses by developing a temporal transformer and a multi-view transformer, which aggregate corresponding features across video frames and multiple views. Similar to Neural Body, Neural Actor \cite{liu2021neuralactor} learns a deformable radiance field with SMPL, while it uses 2D texture maps defined on the body model as the latent codes, which improves the synthesis of pose-dependent dynamic appearance. H-NeRF \cite{h-nerf} proposes to co-learn a radiance field and a signed distance function for rendering and temporally reconstructing dynamic human, conditioned on a geometric prior obtained from an implicit articulated human body model imGHUM. Animatable NeRF \cite{peng2021animatable} introduces a per-frame neural blend weight field to be combined with NeRF, while using human priors from SMPL to regularize the learned blend weight. These methods achieve relatively high-quality performance, but they are primarily intended for multi-view video input. 

Compared to multi-view videos, acquiring monocular videos are more efficient and more accessible for broader applications of free-viewpoint rendering. However, rendering from monocular videos is dramatically more challenging because ill-posed problems are prone to arise due to self-occlusion and inherent depth ambiguity. A-NeRF \cite{su2021anerf} addresses the ill-posed problem by overparameterizing NeRF with skeleton-relative encoding, where its demonstration shows the potential for rendering very articulated motion. HumanNeRF \cite{weng_humannerf_2022_cvpr} considers human motion as a combination of skeleton rigid motion and non-rigid motion that are learned via separate neural networks. HumanNeRF also learns a pose correction network to assist the refinement of the motion field, finally producing high-fidelity rendering results in both a benchmark dataset and in-the-wild videos. NeuMan \cite{jiang2022neuman} proposes to learn a human NeRF and a scene NeRF separately, opening up more opportunities for composition and editing of human dynamic scenes.

The advance of NeRF-based human models promotes the development of dynamic human articulation, animation, and free-viewpoint rendering. However, the aforementioned models all require long training and inference time despite using high-end computational hardware. Accelerating model training and inference is the key to real-time neural rendering applications such as conferencing and gaming.

\subsection{Accelerating NeRF}
NeRF generally requires long per-scene training time and per-image inference time \cite{EfficientNeRF}. For example, on a NVIDIA V100 GPU, vanilla NeRF \cite{mildenhall2020nerf} takes 1-2 days to train a scene with 100 images of 800$\times$800 resolution, and takes 30 seconds to inference an image of the same resolution. The inefficiency hinders its real-world real-time usage.

Besides hierarchical sampling in vanilla NeRF, several methods were developed to accelerate the NeRF training and/or inference. A category of acceleration methods works on modifying the data structures to be more easily accessible. For example, NSVF \cite{liu2020nsvf} organizes a scene into a sparse voxel octree and thus reduces the number of sampling. PlenOctree \cite{yu2021plenoctrees} trains a spherical harmonic NeRF and converts it into a sparse octree representation for increased inference speed. FastNeRF \cite{garbin2021fastnerf} proposes a graphics-inspired factorization approach that enables caching with sparse octree and fast query. These methods only improve the inference speed at the cost of memory, while do not reduce training time. Instant-NGP \cite{mueller2022instant} proposes a learned parametric multi-resolution hash encoding that accelerates both training and inference when applied to NeRF.

Some methods attempt to modify the MLP in NeRF. KiloNeRF \cite{Reiser2021kilonerf} uses thousands of tiny MLPs instead of a deep MLP: it subdivides the scene into thousands of 3D cells with each part represented by a tiny MLP, and thus largely reduces the query time at inference. Instant-NGP \cite{mueller2022instant} demonstrates that integrating hardware-accelerated fully-fused CUDA kernels \cite{mueller2021realtime} into NeRF can increase both training and inference speed.

Some methods optimize the ray marching and volume rendering techniques. In addition to using tiny MLPs, KiloNeRF \cite{Reiser2021kilonerf} employs early ray termination and empty space skipping strategies to improve rendering speed further. Instant-NGP \cite{mueller2022instant} also implements exponential stepping, empty space skipping, and sample compaction. AutoInt \cite{lindell2021autoint} approximates the volume rendering steps and reduces the number of samples for the rendering step, but it compromises the rendering quality. Multiple independent strategies mentioned above can be combined to achieve accelerations of several orders of magnitude. 

The above methods were mainly designed for accelerating static scenes. Little effort has been made for accelerating human neural rendering. More complex than static scenes, a human body contains rigid and non-rigid parts and can perform dynamic movements. The distinct nature requires greater effort in accelerating human rendering, as the existing acceleration methods may not be fully applicable. 

\section{Envisioning the System Pipeline}
\label{sec: design}
\begin{figure*}
    \centering
    \includegraphics[width=1.0\textwidth]{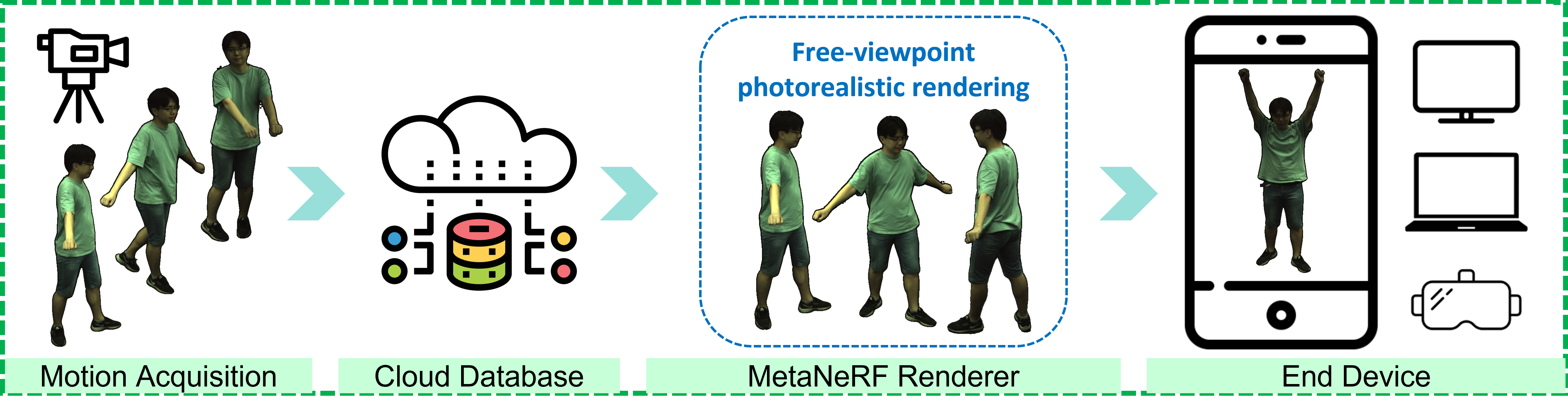}
    \caption{Overview of an envisioned pipeline for extended reality metaverse conferencing using NeRF. The pipeline consists of four building blocks: a motion acquisition module, a cloud database, a MetaNeRF renderer, and a client-end device. The motion acquisition module captures a human's motions using a single camera. The cloud database stores the motion data and transfers them to the MetaNeRF renderer. The renderer combines arts from human animation and NeRF to create photorealistic view synthesis. The client-end device displays synthesized views from the renderer. The pipeline can be flexibly customized according to application needs, with easy extension to create a multi-user conferencing system. The illustration is made with ZJU-MoCap dataset \cite{peng2021neural}.}
    \label{fig:system_new}
\end{figure*}

Existing conferencing systems provide remote communication opportunities, but they cannot fully fulfil the realism needed for an immersive meeting experience. We envision a metaverse conferencing pipeline that addresses the challenges in efficiently achieving a higher level of realism. We explore an accelerated free-viewpoint photorealistic synthesis for rendering human dynamics, with the added simplicity of using a single-camera video acquisition. \cref{fig:system_new} illustrates the design of the system pipeline. Our envisioned pipeline offers a potential solution for low-cost, real-time, and immersive conferencing. 

The system pipeline contains four major components to realize an immersive conferencing experience: a human motion acquisition module, a cloud database, a MetaNeRF renderer, and a client-end device. Unlike most common systems demanding multiple cameras, our motion acquisition module only requires a single camera, which simultaneously improves the hardware and data efficiency. The motion acquisition module captures full-body movement sequences of a dynamic human and uploads the acquired information into the cloud database. The motion information can be in several formats, such as a monocular video or a sequence of video frames. Alternatively, an additional process can be added to extract the key information from the raw data (\eg the appearance profile and 3D skeletal points), which reduces the communication bandwidth. Furthermore, historical data can be leveraged to accelerate data communication in future conferencing events.  

The cloud database stores the motion data. \cref{fig:system_new} presents the database as a central database on a cloud server. It can vary according to the application needs when deployed in the system. For example, the database can be a group of distributed databases on a cloud server or on the edge. Besides the motion data, the database can optionally store a library of background and environment data including lighting that can customize the themes for rendering.   
 
MetaNeRF renderer is the core module for enabling an immersive experience. It combines the arts from human animation and NeRF to create photorealistic view synthesis from any viewpoint. Using the motion data as input, the renderer interprets the human body and camera parameters, and employs neural networks to learn a human motion field. The renderer also leverages an acceleration approach to speed up training and inference. Detailed implementation is discussed in \cref{sec: human-nerf}. The renderer can be flexibly extended for multi-user conferencing and scene relighting. 
 
The client-end device downloads and displays synthesized photorealistic views from MetaNeRF renderer. When compatible and applicable, participants can join metaverse meetings using several types of client-end device, such as smart phone, TV, computer, and AR/VR glasses. Optionally, the client-end device can serve as a computation unit for some efficient computation in the MetaNeRF renderer, or as a private data storage for sensitive personal data. 

With the four components, our envisioned system pipeline can provide an immersive 3D experience for participants joining meetings from distributed locations. The envisioned pipeline allows a high flexibility to customize the components according to the application needs and extend the functionalities for complex scenarios such as theme editing or multi-user conferencing. 

\section{MetaNeRF: Rendering Human Dynamics}
\label{sec: human-nerf}
\begin{figure*}
    \centering
    \includegraphics[width=1.0\textwidth]{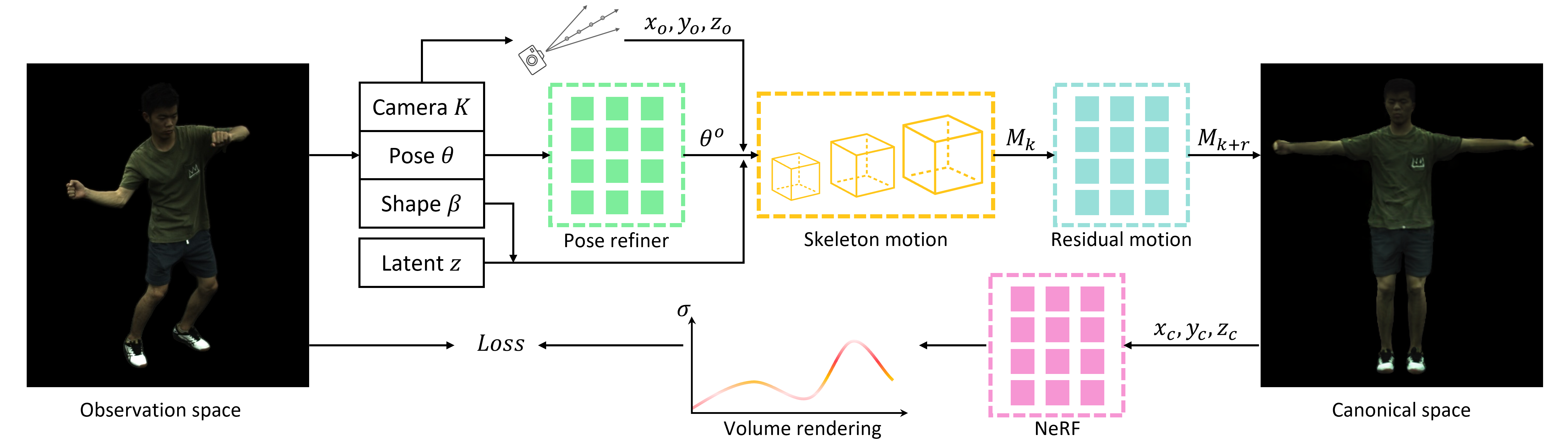}
    \caption{Overview of the framework. Given a sequence of monocular images, we estimate the camera parameters $K$, pose $\theta$ and shape $\beta$ of the human body using an SMPL-based model. The initially estimated pose $\theta$ is gradually refined via a pose refiner network during training. Combining the refined pose $\theta^o$, shape $\beta$, and a latent variable $z$, we learn a motion representation through a skeleton motion network and a residual motion network. We use inverse linear blend skinning to transform the motions in observation space to canonical space. A NeRF-like network is used to learn the color and density maps, which are then volume-rendered into images. We optimize the $Losses$ between the rendered images and the ground truths. This framework is inspired by \cite{weng_humannerf_2022_cvpr}. The networks in the framework are accelerated using fully-fused CUDA kernel \cite{mueller2021realtime}. The illustration is made with ZJU-MoCap dataset \cite{peng2021neural}.}
    \label{fig:human}
\end{figure*}
We explore a more efficient free-viewpoint synthesis algorithm for rendering photorealistic human dynamics from a monocular video. The algorithm demonstrates its effectiveness in realizing an immersive meeting experience and its potential in achieving real-time conferencing. 

We obtain the free-viewpoint synthesis results according to the framework in \cref{fig:human}. Given a sequence of images from a monocular video, we firstly use SPIN (SMPL oPtimization IN the loop) \cite{kolotouros2019spin}, a parametric human body model, to estimate initial camera parameters $K$, body pose $\theta$ and shape $\beta$ of the human body. Compared to the vanilla SMPL model \cite{SMPL:2015} which relies solely on regression, SPIN combines iterative optimization and deep-network regression to estimate human poses more accurately. These SPIN estimations are used as the initial input into the framework, where the pose parameters, particularly, are gradually refined through a pose refiner MLP during training (\cref{subsec: pose}). 

Inspired by previous work \cite{liu2021neuralactor, weng_humannerf_2022_cvpr, peng2022animatable, jiang2022neuman, zheng2022structured}, we represent human motion field $M$ as addition of skeleton-driven motion field $M_{skel}$ and a residual non-rigid motion field $M_{res}$, 
\begin{equation}
    M = M_{\text{skel}} + M_{\text{res}}.
\end{equation}
Specifically, we learn a motion field of the human through two neural networks following \cite{weng_humannerf_2022_cvpr}. One is a convolutional neural network (CNN) that learns the skeleton rigid motion (\cref{subsec: skeletal}), but it is not a full motion representation since it cannot interpret the non-rigid contents. We thus use a MLP to account for the residual non-rigid motion (\cref{subsec: residue}).

\subsection{Pose refiner}
\label{subsec: pose}
Each body pose $\theta$ can be represented as a combination of $K$ joints $J$ and corresponding $K$ joint angles $\Omega = \omega_0, \cdots, \omega_K$. The poses estimated from pre-trained weights of parametric models do not have sufficient accuracy and may lead to pose mismatch. Following \cite{weng_humannerf_2022_cvpr}, we use a MLP to learn an adjustment for a better pose alignment. 

We retain the joints $J$ estimated from images using the SPIN model \cite{kolotouros2019spin}, and optimize an adjustment to each of the $K$ joint angles, $\Delta_\Omega = \Delta \omega_0, \cdots, \Delta \omega_K$. We optimize the network parameters of MLP that provide updates to $\Delta \omega_0, \cdots, \Delta \omega_K$ conditioned on $\omega_0, \cdots, \omega_K$. According to the empirical findings in \cite{weng_humannerf_2022_cvpr},  optimizing the network parameters leads to faster convergence compared to directly optimizing 
$\Delta \omega_0, \cdots, \Delta \omega_K$. 
\begin{equation}
    \Delta_\Omega = \text{MLP}_{\theta}(\Omega)
\end{equation}

We can then update each $\theta$ to $\theta^o$ by corresponding joints, joint angles, and joint angle relatives:
\begin{equation}
    \theta^o = (J, \Delta_\Omega \otimes \Omega).
\end{equation}

\subsection{Human skeleton motion}
\label{subsec: skeletal}
To volumetrically represent the skeleton motion field $M_{\text{skel}}$, researchers typically use either an implicit representation using MLP or an explicit representation using CNN. References \cite{Lombardi_neuralvolume, weng_humannerf_2022_cvpr} discuss the respective advantages and disadvantages of implicit and explicit representations, and adopt the explicit representation with CNN because it is computationally easier and provides smoothness to regularize the optimization. 

Similar to \cite{Lombardi_neuralvolume, weng_humannerf_2022_cvpr, weng2020vid2actor, liu2021neuralactor, peng2022animatable, jiang2022neuman, zheng2022structured}, we model the skeleton motion volume based on an inverse linear blend skinning algorithm that wraps the points in observation space to canonical space (equivalent to warping an observed pose $\theta^o$ to a predefined canonical pose $\theta^c$) in a form as follows:
\begin{equation}
    M_{\text{skel}}(\textbf{x}, \theta^o) = \sum_{i=1}^{K} w_i^{o}(\textbf{x})G_i(\textbf{x}),
\end{equation}
where $w_i^{o}$ is the blend weight for the $i$-th bone in the observation space and $G_i$ is the skeleton motion basis for the $i$-th bone. Practically, $G_i$ is defined as
\begin{equation}
   G_i(\textbf{x}) = R_i\textbf{x} + \textbf{t}_i,
\end{equation}
with $R_i$ and $\textbf{t}_i$ calculated from corresponding body pose $\theta^o$, and $w_i^{o}$ is obtained by first solving the canonical blend weight $w_i^c$ and then deriving from:
\begin{equation}
   w_i^{o}(\textbf{x}) = \frac{w_i^c G_i(\textbf{x})}{\sum_{k=1}^K w_k^c G_k(\textbf{x})}.
   \label{eq:lbs}
\end{equation}

Specifically, we use a CNN to generate a weight volume $W^c(\textbf{x})$, which contains a set of $w_i^c(\textbf{x})$,  from a random constant latent variable, and optimize the network parameters:
\begin{equation}
   W^{c}(\textbf{x}) = \text{CNN}_{\text{skel}}(\textbf{x};\textbf{z}).
\end{equation}


\subsection{Residual motion field}
\label{subsec: residue}
We estimate a residual motion field $M_{\text{res}}$ to account for the non-rigid deformation that is not explained in the skeleton motion field, such as the shifting and folding of clothes. In light of previous works \cite{weng_humannerf_2022_cvpr, peng2022animatable}, we model the residual motion as a pose-dependent deformation field. Specifically, we use a MLP to learn a non-rigid deformation offset conditioned on the skeleton motion field and the body pose:
\begin{equation}
    M_{\text{res}}(\textbf{x}_{\text{skel}},\theta^o) = \text{MLP}_{\text{res}}(\gamma(\textbf{x}_{\text{skel}}), \theta^o),
\end{equation}
where $\textbf{x}_{\text{skel}}$ represents points in skeleton motion field $M_{\text{skel}}$, and $\gamma$ is a positional encoding function.

Adding the non-rigid offset to skeleton motion completes the motion. Points in the motion field can be represented as:
\begin{equation}
    \textbf{x}_{\text{final}} = \textbf{x}_{\text{skel}} + \textbf{x}_{\text{res}},
\end{equation}
where $\textbf{x}_{\text{res}}$ represents points in residual motion field $M_{\text{res}}$.

The residual motion network is not turned on at the early stage of training. This avoids overfitting the residual motion network to the input and undermining the contribution of the skeleton motion. When it joins, we employ a coarse-to-fine manner to the residual motion network with a truncated Hann window applied to the frequency bands of positional encoding \cite{weng_humannerf_2022_cvpr}. At a certain iteration of training, we set it back to full frequency bands of positional encoding.   

\subsection{Learning and representing color and density}
\label{subsec: nerf}
We represent the dynamic human in canonical space as a continuous field, and derive the color $\textbf{c} = (r, g, b)$ and density $\sigma$ using a NeRF-like MLP network:  
\begin{equation}
    \textbf{c}, \sigma = \text{MLP}_{\text{nerf}}(\gamma(\textbf{x}_{\text{final}})),
\end{equation}
where $\gamma$ is a standard positional encoding function. Using the learned $\textbf{c}$ and $\sigma$, we use the volume rendering technique discussed in \cref{subsec: background_nerf} to reconstruct images.

Since the bounding box of a human performer can be estimated from the image, we apply stratified sampling approach \cite{mildenhall2020nerf} inside the bounding box. In addition, we adopt the augmentation method introduced in \cite{weng_humannerf_2022_cvpr} to further improve sampling efficiency. The augmentation method uses the denominator of \cref{eq:lbs} to approximate the likelihood of being part of the human performer, and augment the $(1-\exp{(-\sigma_i\delta_i)})$ in \cref{eq:nerf_3} to be small by multiplying the likelihood when the likelihood is lose to zero.

\subsection{Accelerating training and inference}
\label{subsec: accelerate} 
\begin{figure*}
    \centering
    \includegraphics[width=0.87\textwidth]{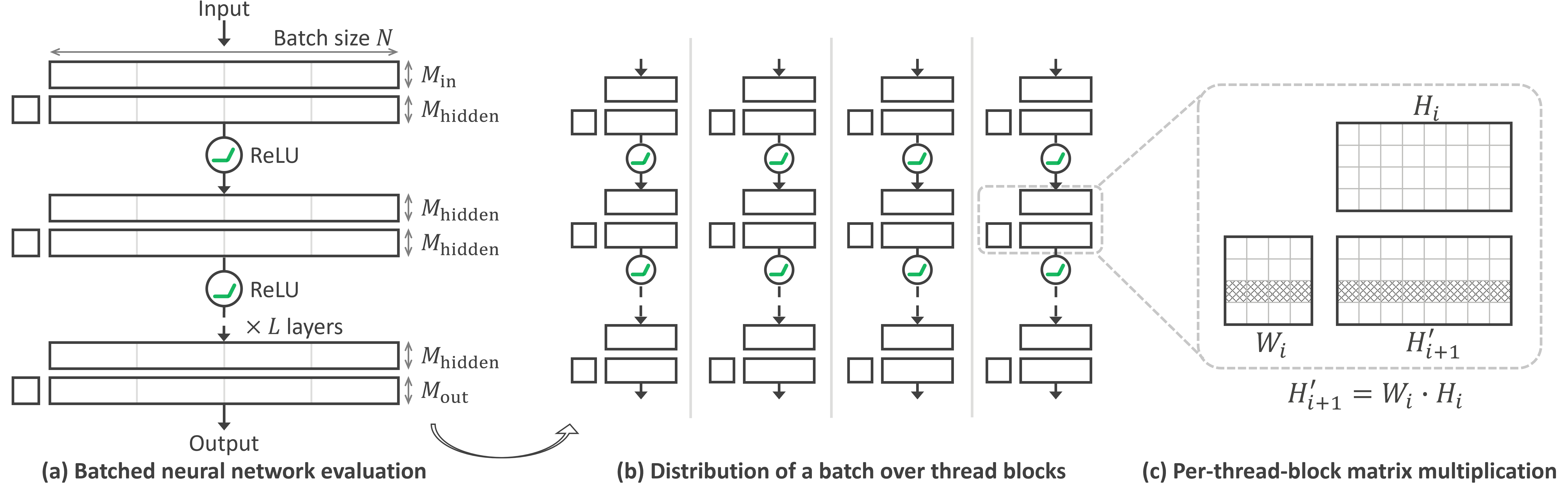}
    \caption{\textbf{(a)} A regular MLP evaluation for a given batch of input vectors corresponds to alternating weight-matrix multiplication and element-wise application of the activation function. \textbf{(b)} A fully fused MLP achieves accelerated performance by parallelizing the workload. It partitions the batch into 128 element wide chunks and processes each chunk by a single thread block. The fully fused MLP is narrow ($M_{hidden}=M_{in}=64\ \text{neurons wide})$, allowing the weight matrices to fit into registers and the intermediate $64 \times 128$ neuron activation to fit into shared memory. \textbf{(c)} Within a matrix multiplication, each thread block transforms the $i$-th layer $H_i$ into the pre-activated next layer $H'_{i+1}$. $H_i$ is diced into $16 \times 16$ elements to match the size of the NVIDIA hardware-accelerated half-precision matrix multiplier TensorCore. Each warp of the thread block computes one $16 \times 128$ block-row (\eg the striped area) of $H'_{i+1}$. The computation is done by first loading the corresponding $16 \times 64$ striped weights in $W_i$ into registers and then multiplying the striped weights by all $64 \times 16$ block-columns of $H_i$. Thus, each thread block loads the weight matrix (\eg $W_i$) from global memory exactly once, while frequent accesses are over $H_i$ located in fast shared memory. This figure is reproduced from \cite{mueller2021realtime}.}
    \label{fig:fullyfused}
\end{figure*}
To increase the training and inference speed, we adopt fully fused neural networks introduced in \cite{mueller2021realtime, mueller2022instant}. As discussed in \cite{mueller2021realtime}, fully fused neural networks take advantage of fully utilizing fast on-chip memory and minimizing traffic to "slow" global memory. We reproduce a figure from \cite{mueller2021realtime} in \cref{fig:fullyfused} to elaborate the mechanism of the fully fused neural networks that leverage the parallelism of modern GPUs. As shown in \cref{fig:fullyfused} (a), given a batch of input vectors, a regular MLP evaluation corresponds to alternating weight-matrix multiplication and element-wise application of the activation function. In contrast, a fully fused MLP in \cref{fig:fullyfused} (b) partitions the given batch of input vectors into block-column segments and processes each segment by a single thread block. The width of fully fused MLP is narrow, enabling the full utilization of fast on-chip memory (\eg registers and shared memory). For a matrix multiplication $H'_{i+1} = W_i \dot H_i$ (\cref{fig:fullyfused} (c)), each warp of the thread block computes one block-row (striped area) of $H'_{i+1}$ by first loading the corresponding striped weights in $W_i$ into registers and then multiplying the striped weights by all block-columns of $H_i$. Thus, each thread block loads the weight matrix (\eg $W_i$) from global memory exactly once, while frequent accesses to $H_i$ are via fast shared memory.  

We implement the MLPs in \cref{sec: human-nerf} as fully fused MLPs using tiny CUDA neural network (tiny-CUDA-nn) framework \cite{tiny-cuda-nn}. We use suggested configurations as in \cite{mueller2021realtime}. 

\section{Experiment}
\subsection{ZJU-MoCap dataset}
We use realistic dataset ZJU-MoCap \cite{peng2021neural} in our experiment for evaluating the system pipeline and algorithm performance in real-world conferencing scenarios. This dataset was captured in the real world by a multi-camera system that contains 20+ synchronized cameras each producing a monocular video. It includes a wide variety of human complex motions, such as warmup, kicking, arm swings, Taichi, and twirling. We select 7 subjects with diverse motions in the experiment. Particularly, we use images captured by camera 1 for training and other camera data for evaluation. We utilize the provided segmentation mask, camera intrinsics and extrinsics, and SMPL parameters of each frame. The resolution of ZJU-MoCaP images is $1024 \text{pi} \times 1024 \text{pi}$.

\subsection{Implementation}
\textbf{Loss function.} We use Mean Squared Error (MSE) and Learned Perceptual Image Patch Similarity (LPIPS) \cite{zhang2018unreasonable} in the loss function. The MSE accounts for the fidelity of pixel-wise appearance, while the LPIPS assesses the perceptual similarity. We optimize the loss between the input frames and the corresponding rendered images with respect to all trainable parameters in networks discussed in \cref{sec: human-nerf}.
\begin{equation}
\label{eq: loss}
    \mathcal{L} = \lambda_{mse} \mathcal{L}_{mse} + \lambda_{lpips} \mathcal{L}_{lpips}
\end{equation}
Following \cite{weng_humannerf_2022_cvpr}, we use $\lambda_{mse} = 0.2, \lambda_{lpips} = 1.0 $ in \cref{eq: loss} and employ VGG as the backbone of LPIPS. We use patch-based ray sampling to accommodate LPIPS loss \cite{weng_humannerf_2022_cvpr, Schwarz2020GRAF}. Specifically, we choose $6$ patches with size $20 \times 20$ on an image, and compare the reconstructed patches against the patches at the same positions on the input image.

\textbf{Training.} We apply Adam optimizer with $\beta_1 = 0.9, \beta_2 = 0.99$. We use a learning rate of $5 \times 10^{-4}$ for the NeRF network, and use $5 \times 10^{-5}$ for other networks. We sample 128 points per ray. For each experiment, training is performed for $150K$ iterations on a single NVIDIA GeForce RTX 3080 Ti GPU. We activate the residual motion network at $10K$ iterations and set it back to full frequency bands of positional encoding at $50K$ iterations. 

\textbf{Evaluation.} To evaluate the rendering performance at unseen camera views, we use additional camera data (\eg camera $2$ to camera $21$) for testing. For each human subject, we sample an image every 30 frames for each camera view from all available cameras, resulting in $350 - 700$ testing images for each subject. Evaluation is conducted on a single NVIDIA GeForce RTX 3080 Ti GPU. We compare our method to HumanNeRF \cite{weng_humannerf_2022_cvpr}, a state-of-the-art method that has relatively higher computational efficiency and rendering quality when compared to some other methods.

\textbf{Metrics.} To quantify the rendering quality, we employ Peak Signal-to-Noise Ratio (PSNR) and Structural Similarity Index (SSIM) in addition to MSE and LPIPS. PSNR is a popular metric for measuring reconstruction fidelity that is affected by corrupting noise. SSIM is a perceptual loss that takes into account luminance, contrast, and structure. These metrics have their own limitations, thus using all four metrics can provide a more comprehensive assessment.

\subsection{Results}
\begin{table*} 
    \centering
    \begin{tabular}{@{} c c c c c c c c c @{}}
    \hline
    \multirow{2}{*}{\textbf{Dataset}}&\multicolumn{2}{c}{\textbf{PSNR} $\uparrow$}&\multicolumn{2}{c}{\textbf{SSIM} $\uparrow$}&\multicolumn{2}{c}{\textbf{LPIPS} $\times1000 \downarrow$} &\multicolumn{2}{c}{\textbf{Time (hour)}  $\downarrow$} \\
    \cline{2-9}
     & \textbf{HumanNeRF}& \textbf{Ours} & \textbf{HumanNeRF}& \textbf{Ours} & \textbf{HumanNeRF}& \textbf{Ours} & \textbf{HumanNeRF}& \textbf{Ours}\\
    \hline
    313 & \textbf{32.20} & 31.93 & \textbf{0.9724} & 0.9696 & \textbf{17.83} & 20.84 & 7.10 & \textbf{3.80} \\
    377 & \textbf{36.30} & 34.46 & 0.9842 & \textbf{0.9847} & 12.23 & \textbf{11.98} & 7.05 & \textbf{3.93} \\
    386 & \textbf{35.29} & 34.56 & 0.9704 & \textbf{0.9764} & 19.02 & \textbf{17.69} & 7.12 & \textbf{3.80} \\
    387 & \textbf{31.64} & 30.99 & \textbf{0.9728} & 0.9681 & \textbf{22.77} & 27.18 & 7.13 & \textbf{4.02} \\
    392 & \textbf{34.90} & 34.18 & \textbf{0.9788} & 0.9778 & \textbf{16.81} & 17.65 & 7.10 & \textbf{3.97} \\
    393 & \textbf{32.24} & 31.93 & \textbf{0.9745} & 0.9729 & \textbf{17.98} & 19.77 & 6.82 & \textbf{3.95} \\
    394 & \textbf{35.13} & 34.11 & \textbf{0.9794} & 0.9759 & \textbf{14.01} & 16.29 & 7.07 & \textbf{3.95} \\
    Avg & \textbf{33.96} & 33.17 & \textbf{0.9761} & 0.9751 & \textbf{17.24} & 18.77 & 7.05 & \textbf{3.92} \\
    $\% \text{Deviation} ^*$ & \multicolumn{2}{c}{-2.33\% } & \multicolumn{2}{c}{-0.10\%} & \multicolumn{2}{c}{ +0.01\%} & \multicolumn{2}{c}{\textbf{-44.5\% (Best: -46.60\%)}}\\

    \hline
    \multicolumn{9}{l}{$^*$ {$\% \text{Deviation} ^*$ = (avg of our method - avg of HumanNeRF) / avg of HumanNeRF * 100\%. Same for other tables. }}
    \end{tabular}
    \caption{Training performance comparison between HumanNeRF \cite{weng_humannerf_2022_cvpr} and our method.}
    \label{tab1}
\end{table*}

\begin{table*} 
    \centering
    \begin{tabular}{@{} c c c c c c c c c @{}}
    \hline
    \multirow{2}{*}{\textbf{Dataset}}&\multicolumn{2}{c}{\textbf{PSNR} $\uparrow$}&\multicolumn{2}{c}{\textbf{SSIM} $\uparrow$}&\multicolumn{2}{c}{\textbf{LPIPS} $\times1000 \downarrow$} &\multicolumn{2}{c}{\textbf{Frame rate (FPS) $\uparrow$}} \\
    \cline{2-9}
     & \textbf{HumanNeRF}& \textbf{Ours} & \textbf{HumanNeRF}& \textbf{Ours} & \textbf{HumanNeRF}& \textbf{Ours} & \textbf{HumanNeRF}& \textbf{Ours} \\
    \hline
    313 & 29.44 & \textbf{29.66} & 0.9676 & \textbf{0.9687} & \textbf{30.33} & 30.42 & 0.353 & \textbf{1.117}\\
    377 & 30.43 & \textbf{30.52} & 0.9754 & \textbf{0.9780} & 24.07 & \textbf{22.22} & 0.413 & \textbf{1.282}\\
    386 & \textbf{33.66} & 33.55 & 0.9743 & \textbf{0.9771} & 31.26 & \textbf{27.84} & 0.452 & \textbf{1.385} \\
    387 & 28.34 & \textbf{28.39} & 0.9641 & \textbf{0.9643} & \textbf{35.72} & 37.45 & 0.380 & \textbf{1.189}\\
    392 & 31.03 & \textbf{31.29} & 0.9702 & \textbf{0.9715} & 34.09 & \textbf{32.45} & 0.380 & \textbf{1.188} \\
    393 & 28.50 & \textbf{28.60} & 0.9605 & \textbf{0.9616} & 37.99 & \textbf{37.90} & 0.355 & \textbf{1.120} \\
    394 & \textbf{29.73} & 29.54 & \textbf{0.9619} & 0.9613  & \textbf{35.87} & 37.16 & 0.377 & \textbf{1.181} \\
    Avg & 30.16 & \textbf{30.22} & \ 0.9677 & \textbf{0.9689} & 32.76 & \textbf{32.20} & 0.387 & \textbf{1.209} \\
    $\% \text{Deviation} ^*$ & \multicolumn{2}{c}{+0.20\%} & \multicolumn{2}{c}{+0.13\%} & \multicolumn{2}{c}{-0.00\%} & \multicolumn{2}{c}{\textbf{+213\% (Best: +216\%)}}\\
    \hline
    \end{tabular}
    \caption{Inference performance comparison between HumanNeRF \cite{weng_humannerf_2022_cvpr} and our method.}
    \label{tab2}
\end{table*}

\begin{figure}
\centering
    \includegraphics[width=0.48\textwidth]{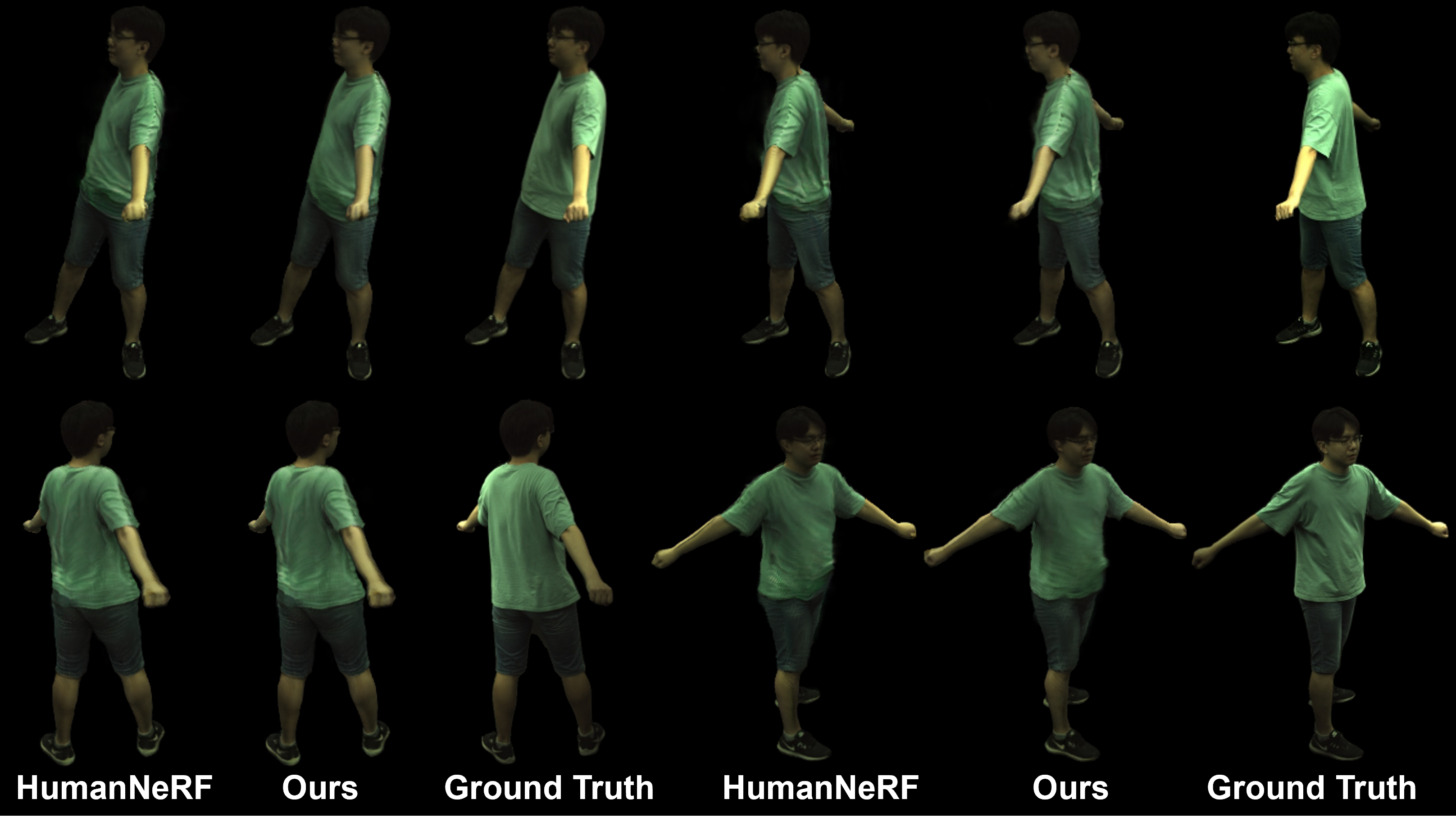}
    \caption{Inference comparison: rendering a human from 4 different viewpoints in the same frame (enlarged in supplementary).} 
    \label{fig:compare1}
\end{figure}

\begin{figure}
\centering
    \includegraphics[width=0.48\textwidth]{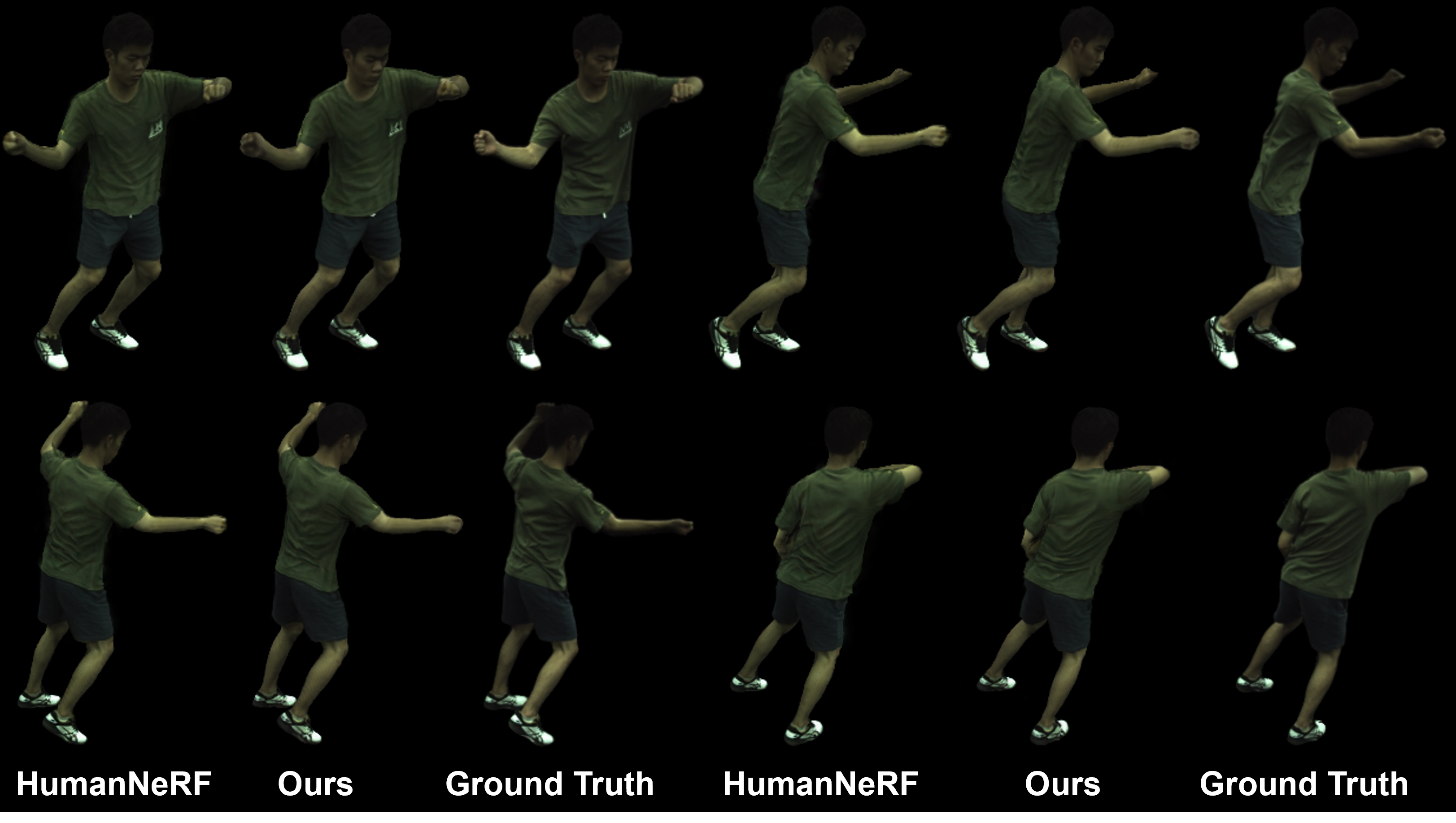}
    \caption{Inference comparison: rendering a human from the same viewpoint at 4 different time frames (enlarged in supplementary).} 
    \label{fig:compare2}
\end{figure}

We compare the performance of our method to that of HumanNeRF \cite{weng_humannerf_2022_cvpr} in terms of rendering quality and time. \Cref{tab1,tab2} present the training performance comparison and inference performance comparison, respectively. \cref{fig:compare1,fig:compare2} showcase visual quality comparisons between our method and HumanNeRF for rendering a human from 4 different viewpoints in the same time frame, and from the same viewpoint at 4 different time frames, respectively.

Quantitatively, the evaluation metrics indicate that our method is significantly more efficient while achieving comparable rendering quality compared to HumanNeRF in both training and inference. On average, compared to HumanNeRF, our method reduces the training time by $44.5\%$ and increases the rendering frame rate by $213\%$. Specifically, our method renders $1024 \text{pi} \times 1024 \text{pi}$ images at a frame rate of $1.209 \ \text{FPS}$ compared to HumanNeRF's $0.387 \ \text{FPS}$. Although there is some variation in the evaluation metrics for image reconstruction between our method and HumanNeRF, the overall difference is insignificant.   

Qualitatively, our method recovers finer and more precise details than HumanNeRF as illustrated in \cref{fig:compare1,fig:compare2}. Our method delivers better contours, while the HumanNeRF has some floating artifacts at the boundaries. However, both our method and HumanNeRF lose some details, such as the clothing wrinkles at the front upper body and the shape of the fingers and hands. Potential improvements can be done by relaxing the constraints imposed in the initial input shape, and by incorporating finer pose parameters.

Overall, our method demonstrates greater efficiency in both training and inference while maintaining comparable quality to HumanNeRF. The performance improvement is mainly a result of fully utilizing on-chip memory and minimizing traffic to slow global memory in the MLP evaluation. However, challenges still remain in realizing our envisioned real-time conferencing, which generally requires a frame rate close to 30 FPS. Unlike static scene rendering, the dynamic motion and pose-dependent deformation of a human body enforce higher acceleration difficulties not only in rendering but also many facets of body and pose reconstruction. While the current work may not fully meet the requirement of our envisioned system, our exploration represents progress in the pursuit of the desired capability. It provides potential insight for future design of such a real-time, lifelike conferencing system.

\section{Conclusion}
We envision a pipeline for a future extended reality conferencing system that offers an immersive and photorealistic experience. At the core of the pipeline, we explore a more efficient free-viewpoint synthesis method with NeRF for rendering human 3D dynamics. Our method achieves state-of-the-art comparable quality and increases training and inference speed by $44.5\%$ and $213\%$ respectively on average. In addition, our method only requires a single-camera motion acquisition, which largely enhances hardware and data efficiency. Our envisioned pipeline provides a design basis for the next generation real-time conferencing systems of low cost, low bandwidth demand, and high accessibility. Future work can focus on improving the rendering speed closer to real-time. Further exploration could be done to extend the system pipeline to enable more complex application scenarios, such as dynamic scene relighting with customized themes and multi-user conferencing that harmonizes real-world people into an extended world.

\clearpage
{\small
\bibliographystyle{ieee_fullname}
\bibliography{paper}
}

\clearpage
\section*{Supplementary Material}
\subsection*{Network architecture}
\begin{figure}[h]
\centering
    \includegraphics[width=0.4\textwidth]{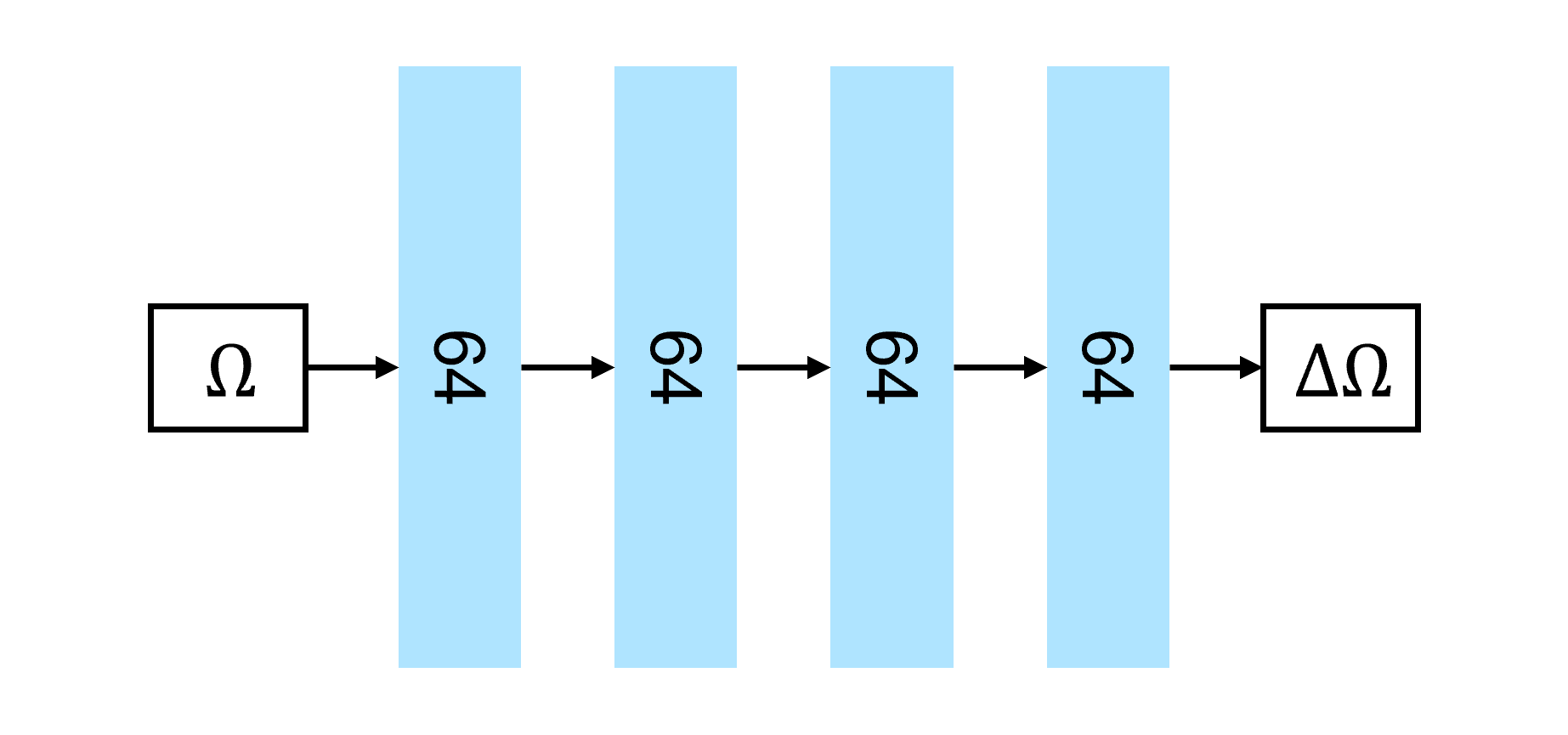}
    \caption{Pose refiner network. Given a body pose $\theta = (J, \Omega)$ in an image, this network takes in joint angles $\Omega$ and outputs joint angle relatives $\Delta \Omega$, which is used to obtain an updated body pose $\theta^o$.}
    \label{fig:pose_refiner}
\end{figure}

\begin{figure}
\centering
    \includegraphics[width=0.45\textwidth]{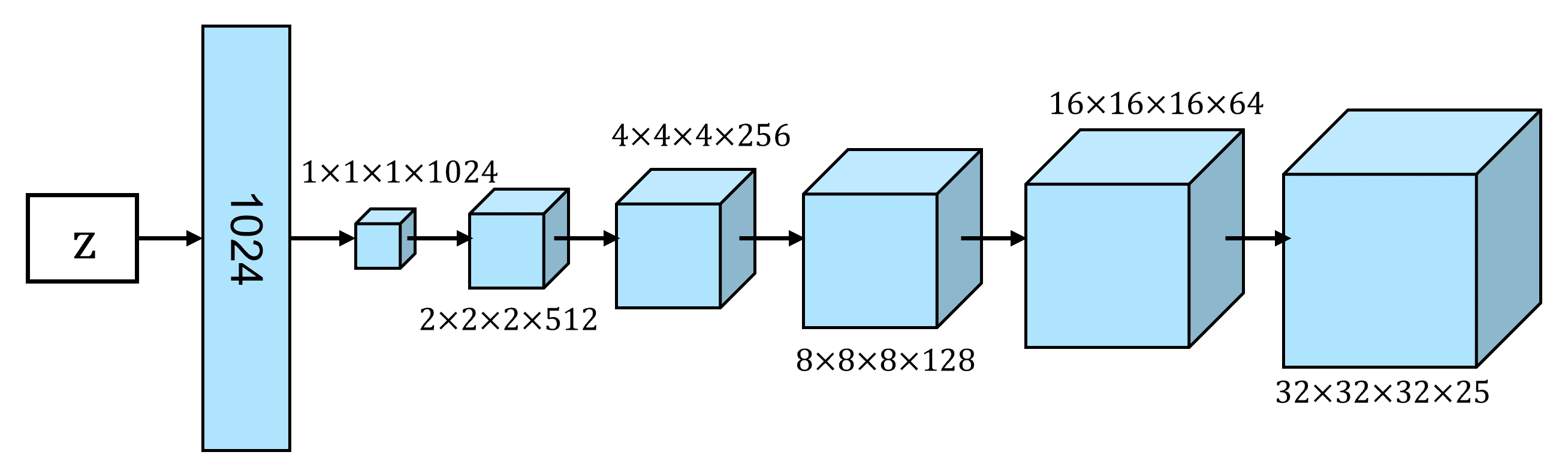}
    \caption{Skeleton motion network. This network generates a weight volume $W^c$ to explicitly represent the skeleton motion field. This network is composed of a fully-connected layer, a tensor reshaping operator, and five 3D transposed convolutions in sequential. It takes in a random constant latent variable $\textbf{z}$ of a size $256$ and outputs a volume of size $32 \times 32 \times 32 \times 25$. The generated weight volume $W^c$ is then used to derive the blend weights $w_i^o$.}
    \label{fig:cnn}
\end{figure}

\begin{figure}
\centering
    \includegraphics[width=0.45\textwidth]{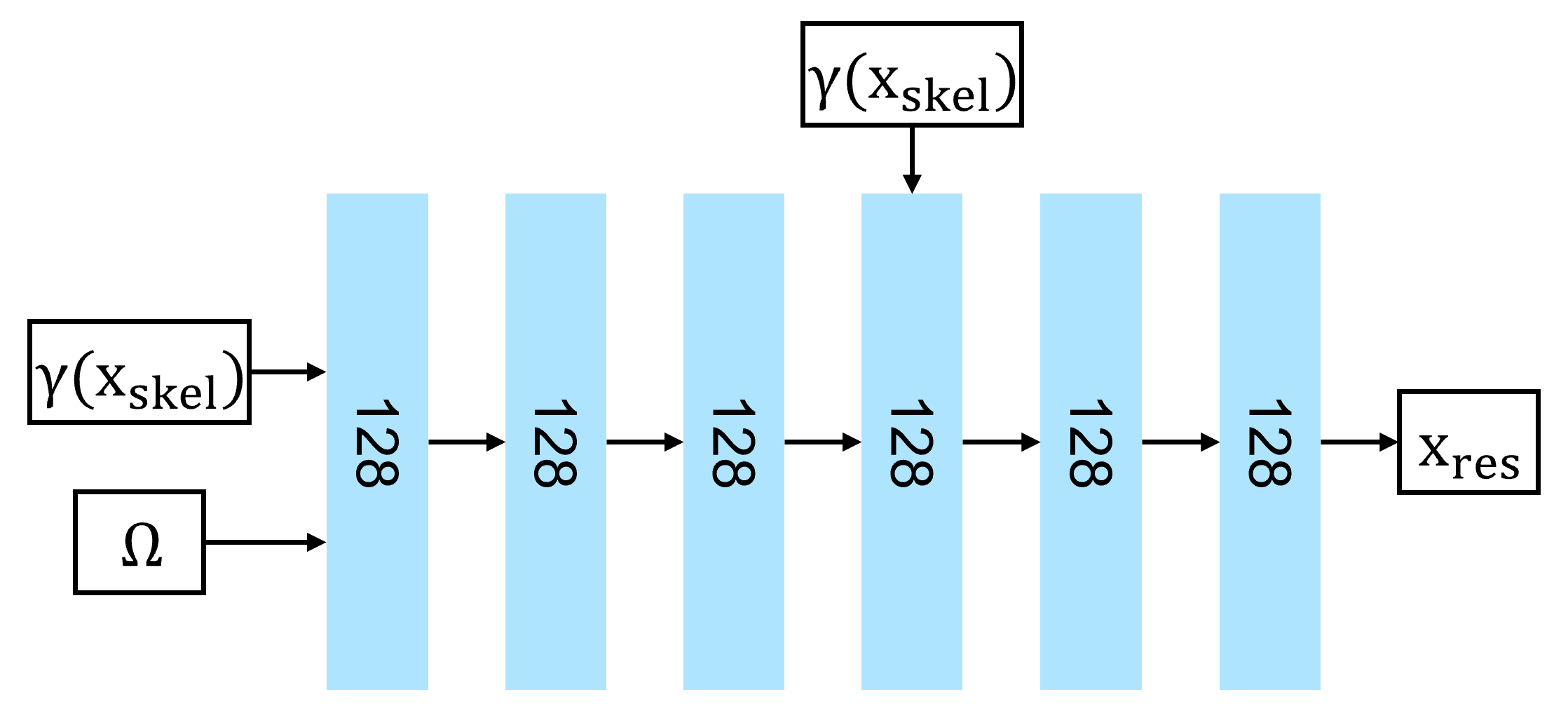}
    \caption{Residual non-rigid motion network. This network is conditioned on the body pose $\theta^o$ and the skeleton motion field $M_{\text{skel}}$. Specifically, it takes in the updated joint angles $\Omega^o$ ($\Omega^o = \Delta_\Omega \otimes \Omega$), and the positional encoding of the points in skeleton motion field $\gamma(\textbf{x}_{\text{skel}})$. At the fourth layer of the network, we use a skip connection for $\gamma(\textbf{x}_{\text{skel}})$. This network produces a residual motion field as an offset $\textbf{x}_{\text{res}}$ to $\textbf{x}_{\text{skel}}$. The addition of $\textbf{x}_{\text{skel}}$ and $\textbf{x}_{\text{res}}$ completes the full motion field. }
    \label{fig:non-rigid_network}
\end{figure}

\begin{figure}
\centering
    \includegraphics[width=0.45\textwidth]{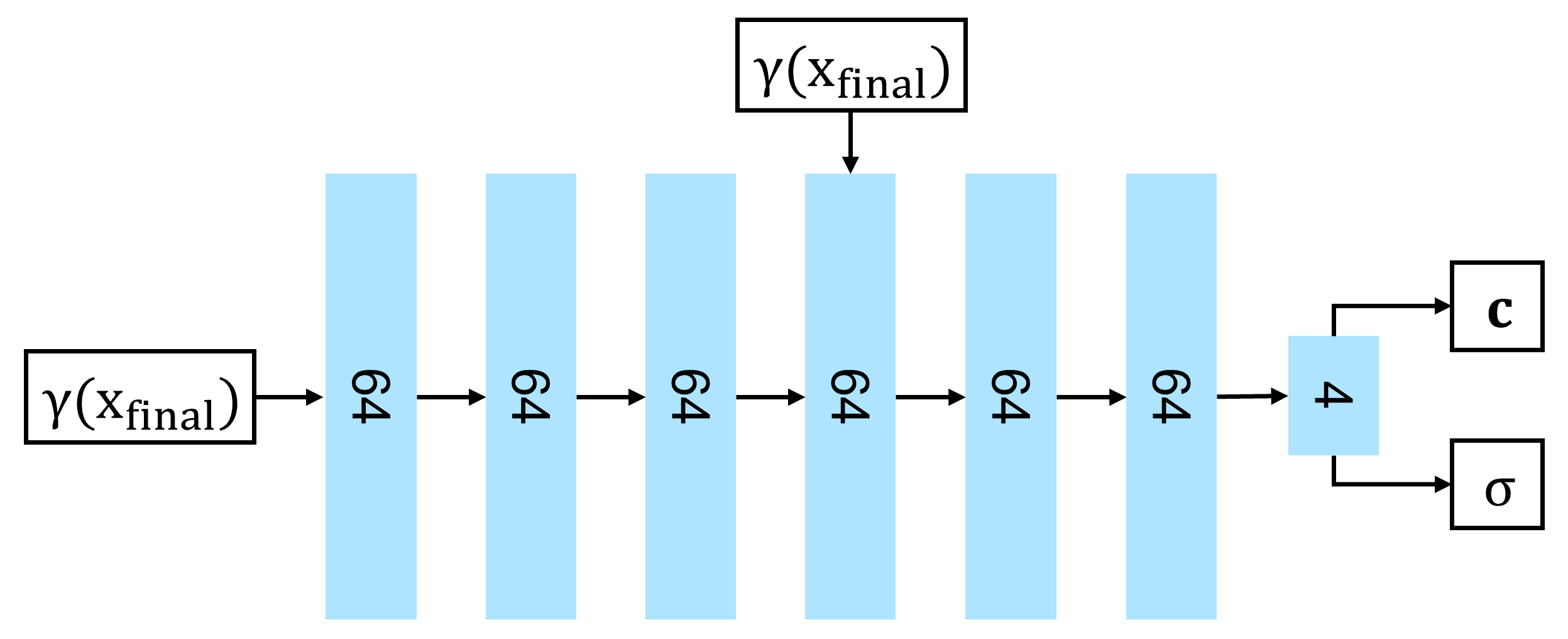}
    \caption{NeRF network for obtaining $\textbf{c}$ and $\sigma$. This network takes in the positional encoding of the points in full motion field $\gamma (\textbf{x}_{\textbf{final}})$, where $\textbf{x}_{\textbf{final}} = \textbf{x}_{\text{skel}} + \textbf{x}_{\text{res}}$. At the fourth layer of the network, we use a skip connection for $\gamma(\textbf{x}_{\text{skel}})$. This network outputs color $\textbf{c}$ and density $\sigma$ for volume rendering.}
    \label{fig:nerf_network}
\end{figure}

\cref{fig:pose_refiner,fig:cnn,fig:non-rigid_network,fig:nerf_network} show the network architecture of pose refiner, skeleton motion, residual non-rigid motion, and NeRF networks. 


\subsection*{Enlarged rendering images}
\begin{figure*}[ht]
\centering
    \includegraphics[width=0.9\textwidth]{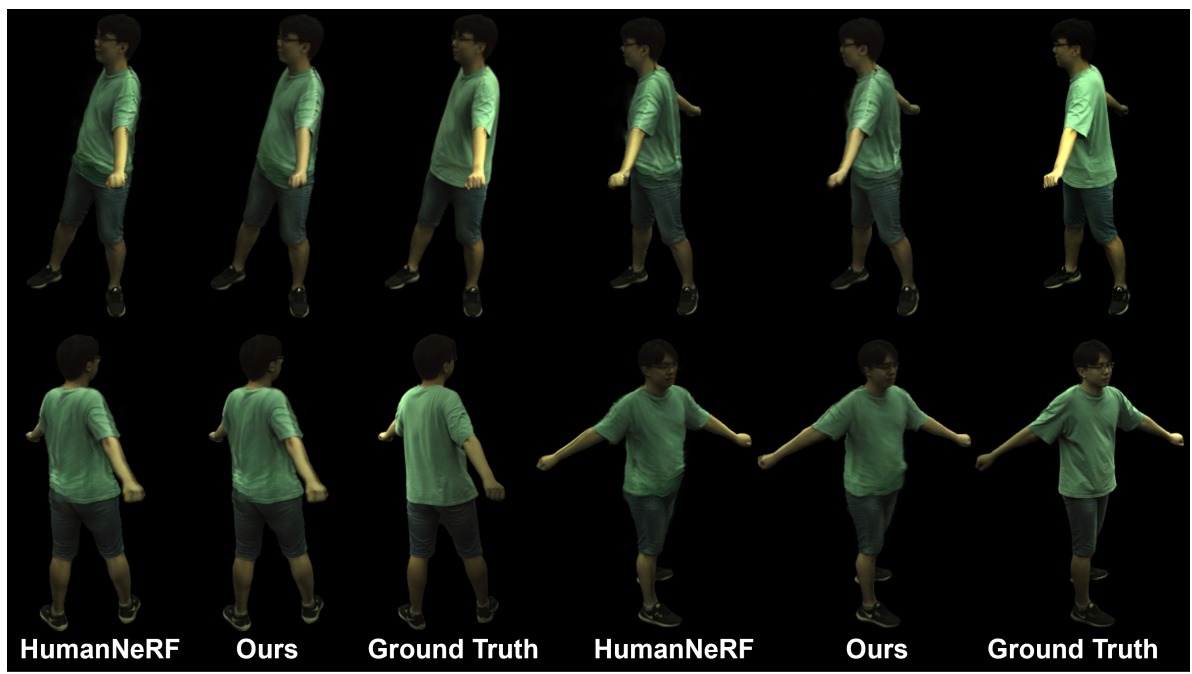}
    \caption{Enlarged \cref{fig:compare1} for detailed inference comparison: rendering a human from 4 different viewpoints in the same time frame.}
    \label{fig:compare4_1_2}
\end{figure*}

\begin{figure*}[ht]
\centering
    \includegraphics[width=0.9\textwidth]{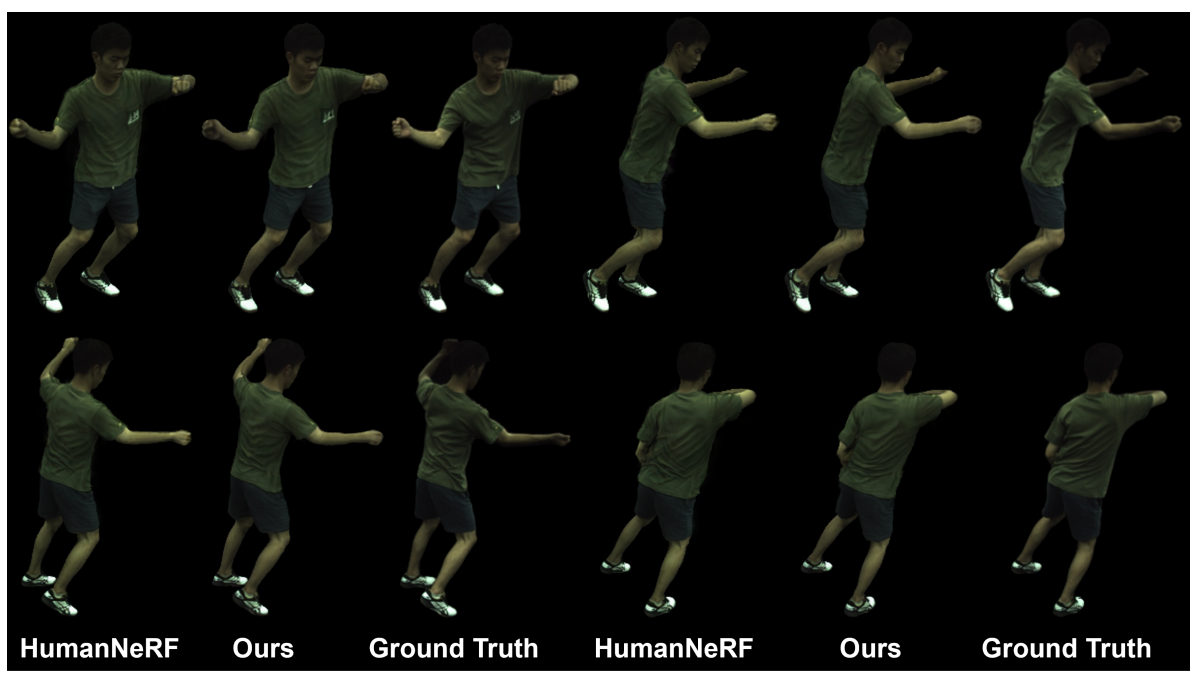}
    \caption{Enlarged \cref{fig:compare2} for detailed inference comparison: rendering a human from the same viewpoint at 4 different time frames.}
    \label{fig:compare4_2_2}
\end{figure*}

\cref{fig:compare4_1_2,fig:compare4_2_2} are enlarged versions of \cref{fig:compare1,fig:compare2}, which showcase a visual quality comparison between our method and HumanNeRF for rendering a human from 4 different viewpoints in the same time frame, and from the same viewpoint at 4 different time frames, respectively.
\end{document}